\RequirePackage{fix-cm}
\documentclass[twocolumn]{svjour3}          
\smartqed  
\usepackage{graphicx}

\usepackage{amsmath}
\usepackage{amssymb}

\usepackage{dcolumn}
\usepackage{bm}
\usepackage{lipsum}
\usepackage{soul}
\usepackage[caption=false]{subfig}

\newcommand{\norm}[1]{\left\lVert#1\right\rVert}
\begin{document}

\title{Forecasting Hamiltonian dynamics without canonical coordinates}



\author{
Anshul Choudhary \and
John F. Lindner  \and
Elliott G. Holliday \and
Scott T. Miller \and
Sudeshna Sinha \and
William L. Ditto
}


\institute{
A. Choudhary \and J. F. Lindner \and E. G. Holliday \and S. T. Miller \and S. Sinha \and W. L. Ditto 
\at Nonlinear Artificial Intelligence Laboratory, Physics Department, North Carolina State University, Raleigh, North Carolina 27607, USA 
\and
J. F. Lindner 
\at Physics Department, The College of Wooster, Wooster, Ohio 44691, USA
\\\email{jlindner@wooster.edu}
\and
S. Sinha
\at Indian Institute of Science Education and Research Mohali, Knowledge City, SAS Nagar, Sector 81, Manauli PO 140 306, Punjab, India
}

\date{Received: date / Accepted: date}

\maketitle

\begin{abstract}
Conventional neural networks are universal function approximators, but because they are unaware of underlying symmetries or physical laws, they may need impractically many training data to approximate nonlinear dynamics. Recently introduced Hamiltonian neural networks can efficiently learn and forecast dynamical systems that conserve energy, but they require special inputs called canonical coordinates, which may be hard to infer from data. Here we significantly expand the scope of such networks by demonstrating a simple way to train them with \textit{any} set of generalised coordinates, including easily observable ones.


\end{abstract}



\section{Introduction}

Artificial neural networks can approximate functions \cite{Cybenko1989,Hornik1991}, model dynamics~\cite{Lusch2018,Jaeger78,Pathak2018,Carroll2019}, elucidate physics \cite{SciNet,AIFeynman,AIPhysicist}, and beat human grand masters at chess and Go~\cite{AlphaZero}. However, they usually require very many high quality training examples, which might be experimentally hard to obtain. Physics-informed neural networks mitigate this problem by encoding some kind of fundamental physics bias, such as time-invariance symmetry. In particular, Hamiltonian neural networks~\cite{HNN,HGN,mattheakis2019,Bertalan2019,Bondesan2019} exploit the symplectic structure of conservative systems to forecast dynamics that mix order and chaos~\cite{choudhary}, even in very high dimensions~\cite{Miller}. They accomplish this by adjusting their weights and biases to minimize a loss function that enforces Hamilton's equations of motion. 

However, Hamiltonian neural networks typically train on canonical variables -- positions and their conjugate momenta -- that might not be known or accessible experimentally. When Greydanus et al.~\cite{HNN} introduced Ham\-iltonian Neural Networks (HHN), they did apply HNN to noncanonical coordinates of the simple pendulum, but their loss function assumed the conjugate momentum equalled the velocity, which is not generally true. Shortly thereafter, Toth et al.~\cite{HGN} introduced Hamiltonian Generative Networks (HGN), which harnessed the Hamiltonian flow without assuming canonical coordinates, but their statistical loss function was a complicated difference of posterior and prior probability distributions. Moreover, HGN learned the dynamics in an abstract space of latent neurons, which is difficult to interpret in terms of canonical coordinates.  

Here we demonstrate a \textit{general} yet \textit{simple} extension of HNN, which uses a neural network preprocessor to train on a set of readily observable generalised coordinates, learn the underlying Hamiltonian, and then accurately forecast the dynamics, even if the training data is contaminated by noise. Example systems include mathematical ecology's famous Lotka-Volterra predator-prey model~\cite{Lotka}, which unexpectedly can be converted into a Hamiltonian system by a nonlinear variable change, an elastic pendulum, whose conjugate momenta are nonlinear combinations of its generalised coordinates, and an even more generic and complicated nonlinear double pendulum.

\section{Neural Network Models} \label{sec:NNModels}

\subsection{Conventional Feed-Forward Neural Network}

In feed-forward artificial neural networks~\cite{Haykin2008}, the activity of neurons in one layer
\begin{equation} \label{eq:sigmoid}
    a_\ell \overset{\text{vec}}{=}  \sigma \left[ w_\ell\, a_{\ell-1}+ b_\ell \right]
\end{equation}
is a vectorised sigmoid function of a linear combination of the activities in the previous layer. The concatenation of such functions eliminates the activities leaving the nonlinear input-output function
\begin{equation} 
   \vec y = \vec f\,[\vec x,w]=\vec{y}_w[\vec x],
\end{equation}
where the weights and biases $w = \{w_\ell,b_\ell\}$. Given many training pairs $\tau = \{\vec x_n, \vec y_n \}$ and a ``loss" function like the mean-square-error
\begin{equation}
    L_w 
    = \norm{\vec y -  \vec f\,[\vec x,w] }^2,
\end{equation}
an optimisation algorithm like stochastic gradient descent finds the best weights and biases
\begin{equation}
    \hat w_\tau= \arg \underset{w}{\min}\, L_w,
\end{equation}
and the trained neural network
\begin{equation}
    \vec y 
    = \vec f\,[\vec x,\hat{w}_\tau]
\end{equation}
approximates the desired function $\vec y\,[ \vec x\,]$.

\subsection{Conventional Neural Network for Dynamics (NN)}

\begin{figure}[tbh!]
    \centering
    \includegraphics[width=1.0\columnwidth]{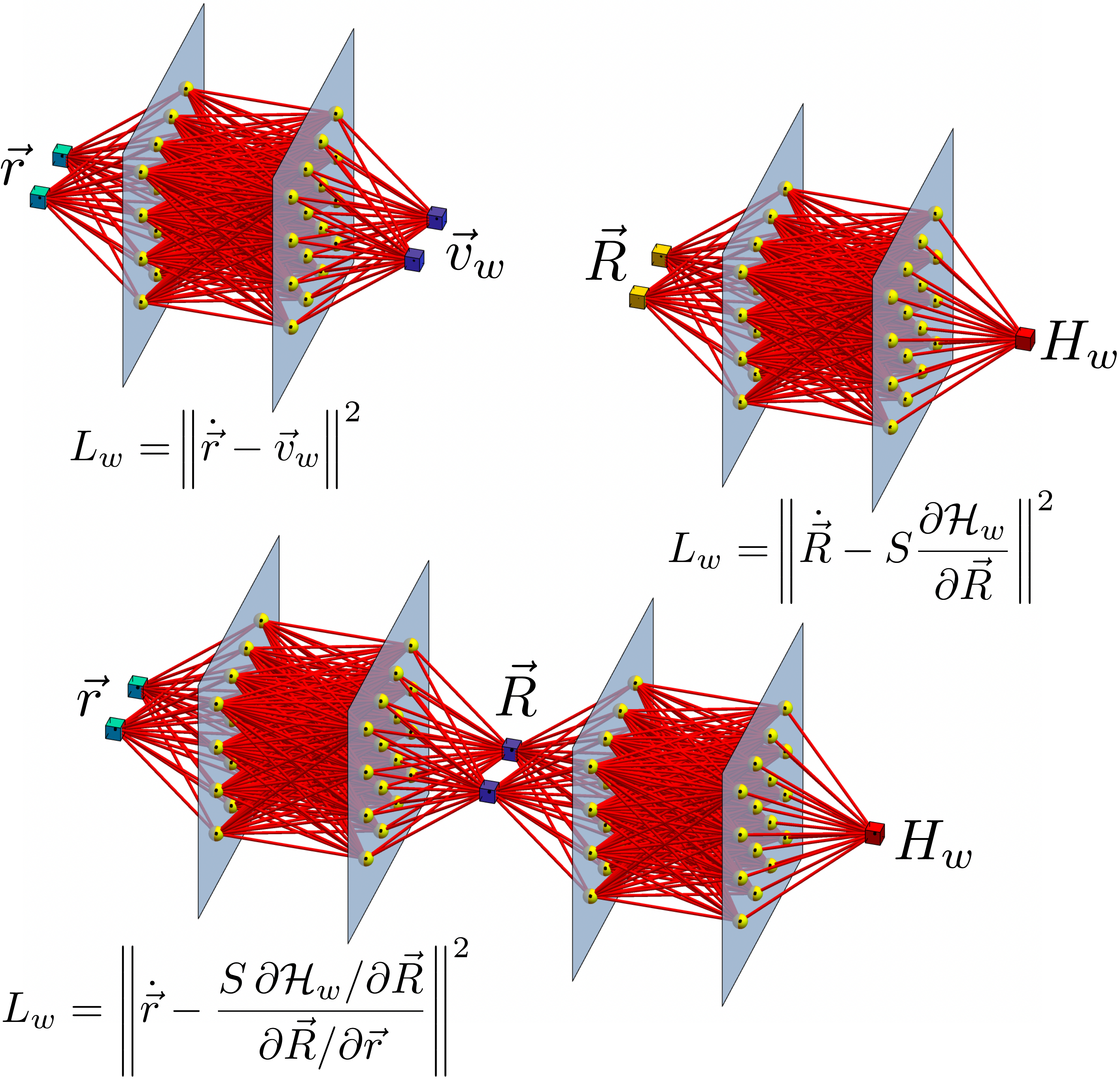}
    \caption{Schematic neural network architectures and loss functions $L$:  NN (top left), HNN (top right), and gHNN (bottom), with generalised coordinates $\smash{\vec r = \{ \vec q,\dot{\vec q}\, \}}$, corresponding velocities $\smash{\vec v = \{ \dot{\vec q},\ddot{\vec q}\, \}}$, canonical coordinates $\smash{\vec R = \{ \vec Q,\vec P \}}$, Hamiltonian $\mathcal{H}$, symplectic matrix $S$, and weights and biases $w$.}
    \label{fig:HNN_schematic}
\end{figure}

To apply a neural network to a dynamical system
\begin{equation} 
   \vec v = \vec v\,[\vec r,w]=\vec{v}_w[\vec r\,],
\end{equation}
intake positions and velocities $\vec r = \{\vec q, \dot{\vec{q}}\, \}$ and output velocities and accelerations $\smash{\dot{\vec r} = \{\dot{\vec q}, \ddot{\vec{q}}\, \}}$, as in Fig.~\ref{fig:HNN_schematic} (top left). With the mean-square-error loss function
\begin{equation}
    L_w = \norm{\dot{\vec r} - \vec v_w [\vec r\,]}^2
\end{equation}
and training pairs $\big\{\vec r, \dot{\vec r}\big\} \rightarrow \big\{\vec q,\dot{\vec q}, \ddot{\vec q}\big\}$, optimise to find the best
\begin{equation}
    \hat w = \arg \underset{w}{\min}\, L_w,
\end{equation}
and use the trained neural network
\begin{equation}
    \dot{\vec{r}} = \vec{v}_{\hat{w}} [\vec r\,]
\end{equation}
to evolve the system forward or backward in time.

\subsection{Hamiltonian Neural Network (HNN)}

To create a Hamiltonian neural network~\cite{HNN,HGN,mattheakis2019,Bertalan2019,Bondesan2019}
\begin{equation} 
  \mathcal{H} 
  = \mathcal{H}\left[\vec R, w\right]
  = \mathcal{H}_w\left[\vec R\,\right],
\end{equation}
intake phase space or canonical coordinates
\begin{equation} 
   \vec R 
   = \left\{ \vec Q,\vec P \right\}
   = \left\{ \vec Q, \frac{\partial \mathcal{L} }{ \partial \dot{\vec Q}} \right\},
\end{equation}
where $\mathcal{L}$ is the Lagrangian, and output a scalar Hamiltonian $\mathcal{H}_w$, as in Fig.~\ref{fig:HNN_schematic} (top right). The Hamiltonian obeys Hamilton's equations
\begin{equation}
\frac{d}{dt}\left[\begin{array}{c} \vec Q \\ \vec P 	\end{array}\right]
= \left[  	\begin{array}{c} +\partial \mathcal{H} / \partial \vec P \\  -\partial \mathcal{H} / \partial \vec Q \end{array}\right]
=    \left[  	\begin{array}{rr} 	0 & \phantom{-}1 \\ -1 & 0 \end{array}\right]
\left[  	\begin{array}{c} \partial \mathcal{H} / \partial \vec Q \\  \partial \mathcal{H} / \partial \vec P \end{array}\right]
\end{equation}
or
\begin{equation} \label{hamEq}
\vec{V}
= \dot{\vec{R}} 
= S \frac{\partial \mathcal{H}}{\partial \vec R}
= S \,\vec\nabla_{\hspace{-0.3em R}} \mathcal{H},
\end{equation}
where $S$ is the symplectic block matrix
\begin{equation}
    S = \left[ \begin{array}{rr} 0 & \phantom{-}1 \\ -1 & 0 \end{array}\right].
\end{equation}
Calculate the gradient $\partial \mathcal{H} / \partial \vec R$ using automatic differentiation~\cite{AutomaticDifferentiation} of the neural network output with respect to its input, and define the mean-square-error loss function
\begin{equation}
    L_w 
    = \norm{\dot{\vec{R}} - \vec V_w[\vec R\,]}^2 
    = \norm{\dot{\vec{R}} -S \frac{\partial \mathcal{H}_w}{\partial \vec R_{\phantom{\hat{w}}}}}^2.
\end{equation}
Optimise over training pairs $\big\{\vec R, \dot{\vec R}\big\} \rightarrow \big\{\vec Q,\vec P, \dot{\vec Q}, \dot{\vec P} \big\}$ to find the best $\hat w$, and use the trained neural network
\begin{equation}
    \dot{\vec{R}} 
    = \vec{V}_{\hat{w}} [\vec R\,]
    = S \frac{\partial \mathcal{H}_{\hat{w}}}{\partial \vec R_{\phantom{\hat{w}}}}
    = S \,\vec\nabla_{\hspace{-0.3em R}} \mathcal{H}_{\hat{w}}
\end{equation}
to evolve the system in time.

\subsection{Generalised Hamiltonian Neural Network (gHNN)}

We want to simply learn a dynamical system's phase space vector field (or differential equations) from the experimentally observed generalised coordinates of sample orbits. However, for most problems, the generalised coordinates are \textit{not} canonical coordinates. Therefore, to leverage the power of HNN, we propose a modified learning architecture where canonical coordinates are effectively learned in an unsupervised way. 

To create a generalised HNN
\begin{equation} 
  \mathcal{H} 
  = \mathcal{H}\left[\vec R \left[ \vec r \, \right], w\right]
  = \mathcal{H}_w\left[\vec R \left[ \vec r \, \right]\right],
\end{equation}
a neural network concatenation intakes generalised positions and velocities $\vec r = \{ \vec x, \dot{\vec{x}} \}$, transforms them to position and conjugate momenta $\smash{ \vec R = \{ \vec Q, \vec P  \} }$ (or some combinations thereof), and outputs a scalar Hamiltonian $\mathcal{H}_w$, as in Fig.~\ref{fig:HNN_schematic} (bottom). The phase space velocities
\begin{equation}
\dot{\vec{R}} 
= \frac{\partial \vec R}{\partial \vec{r}}\frac{d \vec{r}}{dt}
= J \dot{\vec{r}},
\end{equation}
where $J$ is a Jacobian matrix of partial derivatives. Invert to find
\begin{equation}
\vec{v}
= \dot{\vec{r}}
= J^{-1}\dot{\vec{R}}
= J^{-1} S \frac{\partial \mathcal{H}}{\partial \vec R}
\end{equation}
using Hamilton's Eq.~\ref{hamEq}. \ \ Calculate the derivatives  $\partial \mathcal{H}_w / \partial \vec R$ and $\partial \vec R / \partial \vec r$ using automatic differentiation of the neural networks outputs with respect to their inputs, and define the mean-square-error loss function
\begin{align} \label{eq:gHNN1LossFunction}
    L_w 
    &= \norm{\dot{\vec{r}} - \vec v_w[\vec r\,]}^2 \nonumber\\
	&= \norm{ \dot{\vec{r}} - J^{-1} S \vec\nabla_{\hspace{-0.3em R}} \mathcal{H}_w }^2\nonumber\\
	&= \norm{ \dot{\vec{r}} - \frac{S\, \partial \mathcal{H}_w / \partial \vec R}{\partial \vec R / \partial \vec{r}} }^2.
\end{align}
optimise over training pairs $\big\{\vec r, \dot{\vec r}\big\} \rightarrow \big\{\vec q,\dot{\vec q}, \ddot{\vec q}\big\}$ to find the best $\hat w$, and use the trained neural network
\begin{equation}
    \dot{\vec{r}} 
    = \vec{v}_{\hat{w}} [\vec r\,]
    = \frac{S\, \partial \mathcal{H}_{\hat{w}} / \partial \vec R}{\partial \vec R / \partial \vec{r}}
\end{equation}
to evolve the system.

In the special case where the generalised coordinates are the canonical positions, $\vec q = \vec Q$, the Jacobian simplifies to the block matrix
\begin{equation} \label{eq:jac}
    J
    = \frac{ \partial \vec R }{ \partial \vec{r} }
    = \left[ \begin{array}{cc}
       \displaystyle \frac{ \partial \vec Q }{ \partial \vec q}  &\displaystyle \frac{ \partial \vec Q }{ \partial \dot{\vec q}} \\[1.5em]
        \displaystyle\frac{ \partial \vec P }{ \partial \vec q } & \displaystyle\frac{ \partial \vec P }{ \partial \dot{\vec q}  }
    \end{array} \right]
     = \left[ \begin{array}{cc}
        \displaystyle \vphantom{\frac{ \partial \vec Q }{ \partial \vec q}}1  & \displaystyle \vphantom{\frac{ \partial \vec Q }{ \partial \vec q}}0 \\[1.5em]
       \displaystyle \frac{\partial \vec P }{ \partial \vec q}  & \displaystyle\frac{\partial \vec P }{ \partial \dot{\vec q}  }
    \end{array} \right].
\end{equation}

If observed or generalised coordinates $u$ and $v$ relate to an unknown or implicit Hamiltonian with canonical coordinates $Q$ and $P$, then the neural network architecture
\begin{equation} 
  \mathcal{H} 
  = \mathcal{H}\left[\vec R \left[ \vec{\mathcal{O}} \, \right], w\right]
  = \mathcal{H}_w\left[\vec R \left[ \vec{\mathcal{O}} \, \right]\right]
\end{equation}
intakes the observables $u$ and $v$, transforms them to the unknown position and conjugate momenta $Q$ and $P$, and outputs a scalar Hamiltonian $\mathcal{H}_w$. 
In this case, assume a loss function
\begin{equation} \label{eq:gHNN2LossFunction}
	L_w
	= \norm{ \dot{\vec{\mathcal{O}}} -  \dot{\vec{\mathcal{O}}}_w }^2
	= \norm{ \dot{\vec{\mathcal{O}}} - \frac{S\, \partial \mathcal{H}_w / \partial \vec R}{\partial \vec R / \partial \vec{\mathcal{O}}} }^2,
\end{equation}
%
and optimise over training pairs $\smash{\big\{\vec{\mathcal O}, \dot{\vec{\mathcal O}}\big\} \rightarrow \big\{u,v, \dot{u},\dot{v}\big\}}$ to find the best $\hat w$.

\section{Results}

\subsection{Predator-Prey}

The Lotka-Volterra predator-prey model is the ``hydrogen atom" of mathematical ecology~\cite{Lotka}. It is also a nice example of a system that exhibits conservative dynamics without a Hamiltonian or Lagrangian description in its standard variables. Further, since this system arises in the context of population dynamics and has no mechanical analogue, no intuitive equivalent of system kinetic or potential energy exists. So the usual route of constructing a Hamiltonian function from kinetic and potential energy parts is not possible, and consequently the form of the Hamiltonian here is highly non-trivial to guess. Specifically, the coupled nonlinear differential equations governing the population of prey $n_1$ and predator $n_2$ are 
\begin{subequations} \label{eq:LV}
    \begin{align}
        \dot{n}_1 &= +\alpha\, n_1 - \beta\, n_1 n_2, \\
        \dot{n}_2 &= - \gamma\, n_2 +\delta\, n_1 n_2 ,
    \end{align}
\end{subequations}
Notice that neither variable (nor their combinations) can be naturally or readily identified as being coordinate-like or momentum-like. Also, interestingly the combination
\begin{equation} \label{eq:pseudoHam}
    \tilde{\mathcal{H}} = \alpha \log n_2 -\beta n_2 + \gamma \log n_1 - \delta n_1
\end{equation}
is a constant of motion but \textit{not} a Hamiltonian that generates the Eq.~\ref{eq:LV} dynamics. However, the exponential transformation
\begin{subequations}
    \begin{align}
        n_1 &= e^Q, \\
        n_2 &= e^P
    \end{align}
\end{subequations}
implies the coupled system
\begin{subequations} \label{eq:expLV}
    \begin{align}
        \dot Q &= +\alpha - \beta e^P, \\
        \dot P &= -\gamma + \delta e^Q,
    \end{align}
\end{subequations}
where the combination
\begin{align}
    \mathcal{H} =  \alpha P - \beta e^{P} +\gamma Q - \delta e^{Q}
\end{align}
is \textit{both} a constant of the motion \textit{and} a Hamiltonian that generates the Eq.~\ref{eq:expLV} dynamics via
\begin{subequations}
    \begin{align}
        \dot Q &= +\frac{\partial \mathcal{H}}{\partial P}, \\
        \dot P &= -\frac{\partial \mathcal{H}}{\partial Q}.
    \end{align}
\end{subequations}

\begin{figure}[tbh]
    \centering
    \includegraphics[width=1.0\columnwidth]{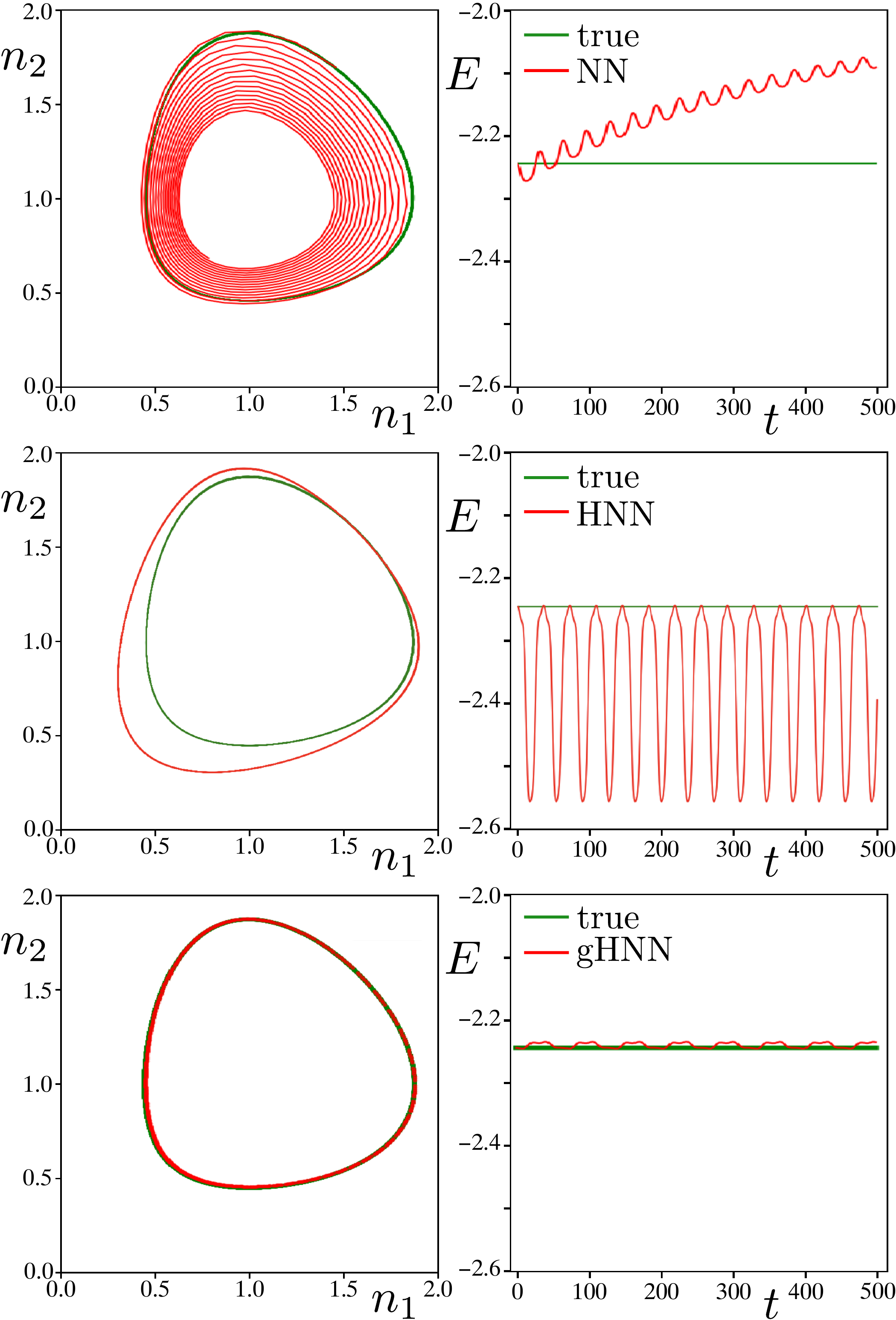}
    \caption{Predator-prey forecasted trajectories $n_2$ versus $n_1$ and corresponding pseudo-energies $E$ versus time $t$ for NN (top), HNN (middle), and gHNN (bottom) trained on generalised coordinates $\{n_1,n_2 \}$ and their derivatives. NN quickly departs the pseudo-energy surface, while HNN is confused by the non-canonical coordinates, but gHNN conserves the pseudo-energy and forecasts well. Parameters are $\alpha=\beta=\gamma=\delta=1$.}
    \label{fig:LV_forecast}
\end{figure}

The learning task is to predict the conservative dynamics by training on the ``ordinary" coordinates $\{n_1,n_2\}$ and their derivatives $\{\dot n_1,\dot n_2\}$, which are the natural observables in the system, without knowing the ``canonical" coordinates $\{Q,P\}$.

The training data consists of 100 trajectories corresponding to different initial conditions, each with a different Eq.~\ref{eq:pseudoHam} pseudo-energy, which demonstrate the famous cycling of predator and prey populations, where the state $\{n_1,n_2\} = \{ \gamma / \delta, \alpha / \beta \}$ is an elliptical fixed point, and the state $\{n_1,n_2\} = \{ 0,0 \}$ is a hyperbolic fixed point. The sampling time $\Delta t = 0.1$ is intentionally large to better approximate real-world data. Implementation details are in Section~\ref{sec:implementDetails}. Parameters are $\alpha = \beta = \delta = \gamma = 1$. 

Each neural network trains identically on the ``ordinary" coordinates $\{n_1,n_2\}$ and their derivatives using the Section~\ref{sec:NNModels} loss functions. Forecasts are made from unseen initial conditions, as in Fig.~\ref{fig:LV_forecast}. NN (top) learns the trajectory for short times, but with no pseudo-energy constraint, it gradually diverges from the true dynamics. HNN (middle) learns some other trajectory and does not preserve the pseudo-energy. With its concatenated neural networks, gHNN (bottom) learns both the trajectory and the pseudo-energy, because its first neural network approximates the relationship between the ``ordinary" and canonical coordinates.

\subsection{Elastic Pendulum}

The elastic pendulum is a simple mechanical systems that exhibits fascinating behaviour. It is one of the simplest systems whose canonical momenta are nontrivial combinations of its coordinates. If the pendulum has length $\ell = r$ and is at an angle $\theta$ from downward, then the pendulum mass $m$ is at position
\begin{equation}
    \vec r = \{x,y\} =  \ell \{ \sin \theta, - \cos \theta \}
\end{equation}
moving with velocity
\begin{equation}
    \vec v 
    = \dot{\vec r} 
    =\dot{ \ell} \{ \sin \theta, - \cos \theta \}+ \ell \{ \cos \theta, \sin \theta \}\dot \theta.
\end{equation}
The Lagrangian
\begin{align}
    \mathcal{L} 
    &= \mathcal{T} - \mathcal{V} \nonumber\\
    &= \frac{1}{2} m v^2 + m g \ell \cos \theta - \frac{1}{2}k(\ell-\ell_0)^2,
\end{align}
where $m$ is the mass, $k$ is the stiffness, $\ell_0$ is the equilibrium length, and $\vec g$ is the gravitational field. The conjugate momenta
\begin{subequations}
    \begin{align}
        p_\ell &= \frac{\partial \mathcal{L}}{\partial \dot{\ell}} = m \dot \ell, \\
        p_\theta &= \frac{\partial \mathcal{L}}{\partial \dot{\theta}} = m \ell^2 \dot \theta,
        \end{align}
\end{subequations}
where $p_\theta$ is \textit{not} simply mass times velocity. 

\

\begin{figure}[tbh]
    \centering
    \includegraphics[width=1.0\columnwidth]{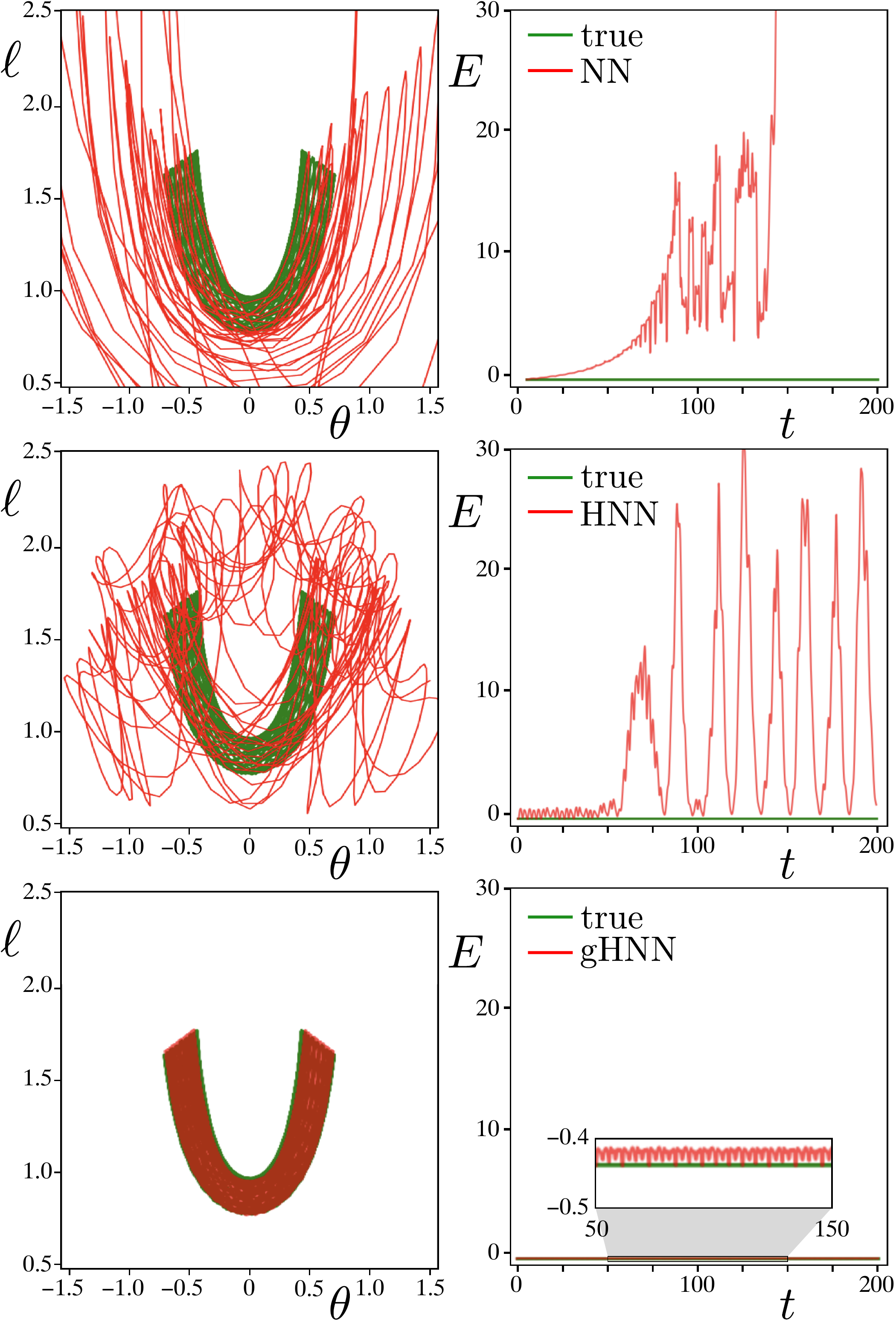}
    \caption{Elastic pendulum forecasted trajectories $\ell$ versus $\theta$ and corresponding energies $E$ versus time $t$ for NN (top), HNN (middle), and gHNN (bottom) trained on generalised coordinates $\{\ell,\theta\}$ and their derivatives. NN diverges from the energy surface, while HNN is confused by the non-canonical coordinates, but gHNN respects the energy and forecasts well, with relatively negligible error. Parameters are $m = g = \ell_0 = 1$ and $k = 4$.}
    \label{fig:EP_forecast}
\end{figure}

The learning task is to predict the conservative dynamics by training on the generalised coordinates $\{\ell,\theta\}$ and their derivatives $\{\dot \ell, \dot \theta, \ddot \ell, \ddot \theta\}$ without knowing the canonical coordinates $\{\ell,\theta, p_\ell, p_\theta \}$. Parameters are $m = g = \ell_0 = 1$ and $k = 4$. The training data consists of 100 trajectories corresponding to different initial conditions, each with a different energy, again coarsely sampled.

Just as we use stochastic gradient descent to optimise our weights and biases, we also vary the \textit{initial} weights and biases \textit{and} our training or hyperparameters to seek the deepest loss minimum in the very high dimensional landscape of possibilities. One strategy is to repeat the computation multiple times from different starts, disregard the outliers and the occasional algorithmic errors (such as not-a-number NaNs or singular value decomposition failures, which might occur in computing the inverse of the Eq.~\ref{eq:jac} Jacobian) and average the remaining results~\cite{Miller}.

Each neural network trains identically on generalised coordinates $\{\ell, \theta \}$ and their derivatives using the Section~\ref{sec:NNModels} loss functions. Forecasts are made from unseen initial conditions, as in Fig.~\ref{fig:EP_forecast}. NN (top) learns the trajectory for short times, but with no energy constraint, it dramatically diverges from the true dynamics. HNN (middle) seems confused by the generalised coordinates when it expect canonical coordinates, and it neither conserves energy nor learns the trajectory, even qualitatively. With its concatenated neural networks, gHNN (bottom) learns the trajectory from the generalised coordinates even for long times, because its first neural network approximates the relationship between the generalised and canonical coordinates.

\subsection{Double Pendulum}

As a final and most challenging example, consider librations of a double pendulum. 
This is a classic chaos demonstrator, \textit{both} of whose canonical momenta are nontrivial combinations of its coordinates. If the the pendulum lengths $\ell_1$ and $\ell_2$ are at angles $\theta_1$ and $\theta_2$ from downward, then the masses $m_1$ and $m_2$ are at positions
\begin{subequations}
    \begin{align}
        \vec r_1 &= \{x_1,y_1\} =  \ell_1 \{ \sin \theta_1, - \cos \theta_1 \}, \\
        \vec r_2 &= \{x_2,y_2\} =  \ell_2 \{ \sin \theta_2, - \cos \theta_2 \} + \vec r_1 
    \end{align}
\end{subequations}
moving with linear velocities
\begin{subequations}
    \begin{align}
        \vec v_1 &= \dot{\vec r}_1 = \ell_1 \{ \cos \theta_1, \sin \theta_1 \} \dot \theta_1, \\
        \vec v_2 &= \dot{\vec r}_2 = \ell_2 \{ \cos \theta_2, \sin \theta_2 \} \dot \theta_2 + \vec v_1.
    \end{align}
\end{subequations}
%
%
The Lagrangian
\begin{align}
    \mathcal{L} 
    &= \mathcal{T} - \mathcal{V} \nonumber\\
    &= \frac{1}{2} m_1 v_1^2 +\frac{1}{2} m_2 v_2^2 + m_1 g y_1 + m_2 g y_2 \nonumber\\
    &= \frac{1}{2}(m_1 + m_2)\ell_1^2 \dot \theta_1^2 
    + \frac{1}{2} m_2\ell_2^2  \dot \theta_2^2
    + m_2 \ell_1 \ell_2 \dot \theta_1 \dot \theta_2 \cos[\theta_1 - \theta_2] \nonumber\\
    &+ (m_1 + m_2) g\ell_1  \cos \theta_1
    + m_2 g \ell_2 \cos \theta_2,
\end{align}
where $\vec g$ is the gravitational field. The conjugate momenta
\begin{subequations}
    \begin{align}
        p_1 &= \frac{\partial \mathcal{L}}{\partial \dot{\theta_1}} 
        = (m_1 + m_2) \ell_1^2 \dot \theta_1 + m_2 \ell_1 \ell_2 \dot \theta_2 \cos[\theta_1 - \theta_2], \\
        p_2 &= \frac{\partial \mathcal{L}}{\partial \dot{\theta_2}} 
        = m_2 \ell_2^2 \dot \theta_2 + m_2 \ell_1 \ell_2 \dot \theta_1 \cos[\theta_1 - \theta_2],
    \end{align}
\end{subequations}
where \textit{neither} $p_1$ nor $p_2$ is simply mass times velocity.

The learning task is to predict the conservative dynamics by training on the generalised coordinates $\{\theta_1, \theta_2\}$ and their derivatives $\{\dot \theta_1, \dot \theta_2, \ddot \theta_1, \ddot \theta_2\}$ without knowing the canonical coordinates $\{\theta_1, \theta_2, p_1, p_2\}$. Parameters are $m_1 = m_2 = \ell_1 = \ell_2 = g =  1$. The training data consists of 100 trajectories corresponding to different initial conditions.

\begin{figure}[bt!]
    \centering
    \includegraphics[width=1.0\columnwidth]{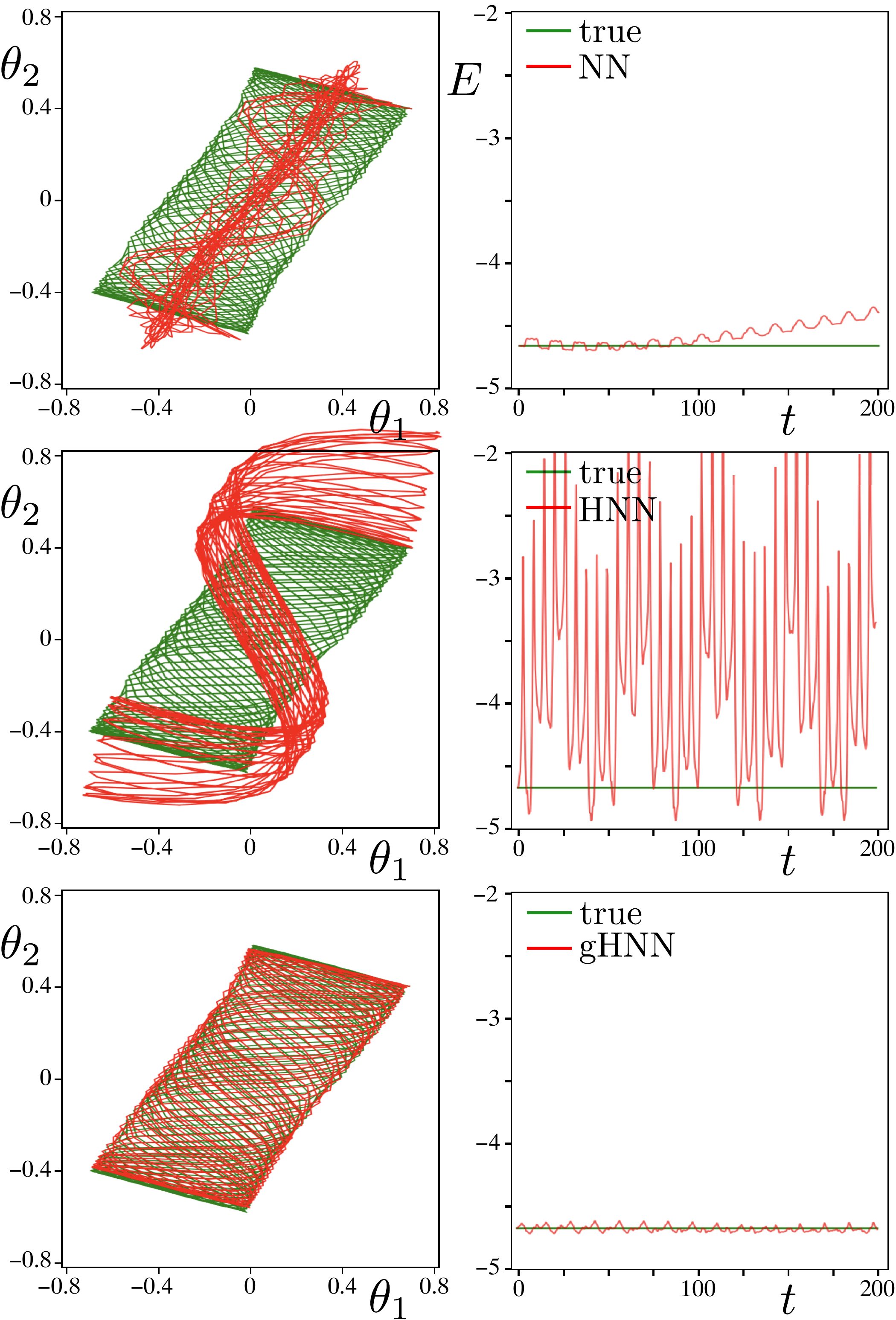}
    \caption{Double pendulum forecasted trajectories $\theta_2$ versus $\theta_1$ and corresponding energies $E$ versus time $t$ for NN (top), HNN (middle), and gHNN (bottom) trained on generalised coordinates $\{\theta_1,\theta_2 \}$ and their derivatives. NN wanders from the energy surface, while HNN is confused by the non-canonical coordinates, but gHNN respects the energy and forecasts well. Parameters are $m_1 = m_2 = \ell_1 = \ell_2 = g =  1$.}
    \label{fig:DoublePendulumForecast}
\end{figure}

Each neural network trains identically on generalised coordinates $\{\theta_1, \theta_2\}$ and their derivatives using the Section~\ref{sec:NNModels} loss functions. Forecasts are made from unseen initial conditions, as in Fig.~\ref{fig:DoublePendulumForecast}. NN (top) learns the trajectory for short times, but with no energy constraint, it gradually diverges from the true dynamics. HNN (middle) is confused by the non-canonical coordinates, and fails dramatically: it tries to learn \textit{some} Hamiltonian flow, but not the right one. If HNN tries to force a Hamiltonian structure where none exists (that is, tries to find area-preserving flows in the space of generalised coordinates), it can fail even more than conventional NN. gHNN (bottom) learns the trajectory from the generalised coordinates even for long times, because its first neural network approximates the relationship between the generalised and canonical coordinates. gHNN performs well even when the training data is contaminated by modest amounts of additive noise (generated by pseudo-random numbers).

\section{Conclusions}

The analysis of classical systems using canonical coordinates, whose time dependence is given by Hamilton's equations of motion, offers one of the most powerful frameworks for understanding classical mechanics across spatial and temporal scales. While it may be easier to formulate the forces and obtain data for a physical system in one set of generalised coordinates, inspecting the motion in the abstract phase space of canonical coordinates helps us discover conserved quantities and fundamental symmetries underlying the motion. Though the knowledge of canonical coordinates is crucial for the formulation of Hamiltonian dynamics, it may not be easy or straight-forward to obtain real-world data in terms of these variables. In fact, finding combinations of generalised coordinates and generalised velocities that yield canonical coordinates and momenta is one of the hard problems of classical physics. 

We sought a Hamiltonian neural network formalism unconstrained by the necessity of inputting data in canonical coordinates. As benchmark and proof-of-principle, here we introduce a simple learning architecture that can forecast Hamiltonian dynamics without explicit knowledge of canonical coordinates. Unfettered by the difficult task of finding appropriate canonical coordinates, our physics-informed neural network takes generalised coordinates as input. \textit{Any} set of coordinates that are amenable to observations will suffice. Coordinates may even derive from video tracking, with velocities and accelerations obtained either by finite differencing or by variational techniques that are robust with respect to noise~\cite{Chartrand}.

We demonstrate the success of this generalised coor\-dinate-based HNN on three nonlinear dynamical systems: a classic population dynamics model and prototypical mechanical models that exhibit both order and chaos. The simplicity and generality of gHNN expands the scope of physics-informed machine learning.

 \section{Implementation Details} \label{sec:implementDetails}

 In our three examples, with phase space dimensions of $d=2$ or $d=4$, NN has $d$ inputs, 2 layers of 50 neurons, and $d$ outputs for a $d$:50:50:$d$ architecture. HNN has $d$ inputs, 2 layers of 200 neurons, and 1 output for a $d$:200:200:1 architecture. gHNN is the concatenation of NN and HNN for a $d$:50:50:$d$:200:200:1 architecture. All neurons use hyperbolic-tangent sigmoids in Eq.~\ref{eq:sigmoid}. The neural networks run on desktop computers and are implemented using the PyTorch library.

\begin{acknowledgements}
This research was supported by ONR grant N00014-16-1-3066, a gift from United Therapeutics, and support from Aeris Rising, LLC. J.F.L. thanks The College of Wooster for making possible his sabbatical at NCSU. S.S. acknowledges support from the J.C. Bose National Fellowship (Grant No. SB/S2/JCB-013/2015).
\end{acknowledgements}

\section*{Conflict of interest}
The authors declare that they have no conflict of interest.

\bibliographystyle{spmpsci}      

\bibliography{iHNN}


\end{document}